\documentclass[11pt]{article}
\pdfoutput=1
\usepackage{smile}

\usepackage{amsmath,amssymb}
\usepackage{bm}
\usepackage{natbib}
\usepackage[usenames]{color}
\usepackage{amsthm}

\usepackage{multirow} 

\usepackage[colorlinks,
linkcolor=red,
anchorcolor=blue,
citecolor=blue
]{hyperref}

\usepackage{fullpage}
\usepackage[protrusion=true,expansion=true]{microtype}
\usepackage{comment}

\usepackage{url}
\usepackage{xcolor}%%table

\usepackage{paralist, tabularx}
\newcommand{\ignore}[1]{}

\begin{document}
\title{\huge A Frank-Wolfe Framework for Efficient and Effective Adversarial Attacks}

\author
{
    Jinghui Chen\thanks{Department of Computer Science, University of Virginia, Charlottesville, VA 22904, USA; e-mail: {\tt jc4zg@virginia.edu}} 
    ~~~~and~~~~Dongruo Zhou\thanks{Department of Computer Science, University of California, Los Angeles, CA 90095, USA; e-mail:  {\tt drzhou@cs.ucla.edu}}
    ~~~~and~~~~Jinfeng Yi\thanks{JD AI Research, Beijing 100101, China; e-mail: {\tt yijinfeng@jd.com}}
	~~~~and~~~~Quanquan Gu\thanks{Department of Computer Science, University of California, Los Angeles, CA 90095, USA; e-mail:  {\tt qgu@cs.ucla.edu}}
}

\date{}
 
\maketitle

\begin{abstract}
Depending on how much information an adversary can access to, adversarial attacks can be classified as white-box attack and black-box attack. 
For white-box attack, optimization-based attack algorithms such as projected gradient descent (PGD) can achieve relatively high attack success rates within moderate iterates. However, they tend to generate adversarial examples near or upon the boundary of the perturbation set, resulting in large distortion. Furthermore, their corresponding black-box attack algorithms also suffer from high query complexities, thereby limiting their practical usefulness. 
In this paper, we focus on the problem of developing efficient and effective optimization-based adversarial attack algorithms. In particular, we propose a novel adversarial attack framework for both white-box and black-box settings based on a variant of Frank-Wolfe algorithm. We show in theory that the proposed attack algorithms are efficient with an $O(1/\sqrt{T})$ convergence rate.
The empirical results of attacking the ImageNet and MNIST datasets
also verify the efficiency and effectiveness of the proposed algorithms. 
More specifically, our proposed algorithms attain the best attack performances in both white-box and black-box attacks among all baselines, and are more time and query efficient than the state-of-the-art.
\end{abstract}

\section{Introduction}\label{sec:intro}
Deep Neural Networks (DNNs) have made many breakthroughs in different areas of artificial intelligence such as image classification \citep{krizhevsky2012imagenet,he2016deep}, object detection \citep{ren2015faster,girshick2015fast}, and speech recognition \citep{mohamed2012acoustic,bahdanau2016end}. However, recent studies show that deep neural networks are vulnerable to adversarial examples \citep{szegedy2013intriguing,goodfellow6572explaining} -- a tiny  perturbation on an image that is almost invisible to human eyes could mislead a well-trained image classifier towards misclassification. Soon later this is proved to be not a coincidence in image classification: similar phenomena have been observed in other problems such as speech recognition \citep{carlini2016hidden}, visual QA~\citep{xu2017can}, image captioning~\citep{chen2017show}, machine translation~\citep{cheng2018seq2sick}, reinforcement learning \citep{pattanaik2018robust}, and even on systems that operate in the physical world \citep{kurakin2016adversarial}. 

Depending on how much information an adversary can access to, adversarial attacks can be classified into two classes:  white-box attack~\citep{szegedy2013intriguing,goodfellow6572explaining} and black-box attack~\citep{papernot2016transferability,chen2017zoo}. In the white-box setting, the adversary has full access to the target model, while in the black-box setting, the adversary can only access the input and output of the target model but not its internal configurations. %To conduct adversarial attacks, various types of attack algorithms were proposed in the literature~\citep{goodfellow6572explaining,moosavi2016deepfool,baluja2017adversarial,xiao2018generating,su2017one,papernot2016limitations,moosavi2017universal, papernot2016transferability,szegedy2013intriguing,kurakin2016adversarial, carlini2017towards,chen2017ead, chen2017zoo,ilyas2018black,cheng2018query}. 

Several optimization-based methods have been proposed for the white-box attack.
One of the first successful attempt is the FGSM method \citep{goodfellow6572explaining}, which works by linearizing the network loss function. CW method \citep{carlini2017towards} further improves the attack effectiveness by designing a regularized loss function based
on the logit-layer output of the network and optimizing the loss by Adam \citep{kingma2014adam}. Even though CW largely improves the effectiveness, it requires a large number of gradient iterations to optimize the distortion of the adversarial examples.
Iterative gradient (steepest) descent based methods such as PGD \citep{madry2017towards} and I-FGSM \citep{kurakin2016adversarial} can achieve relatively high attack success rates within a moderate number of iterations. However, they tend to generate adversarial examples near or upon the boundary of the perturbation set, due to the projection nature of the algorithm. This leads to large distortion in the resulting adversarial examples.
%On the other hand, the efficiency of the current white-box attack algorithms is still far from optimal, leading to high computational burden on large-scale tasks. For example, in adversarial training \citep{madry2017towards}, adversarial attacks on the whole batch of the training samples are needed before each weight update, causing significantly longer model training time compared with natural training procedure,  
%which is far from satisfactory.
% Among the approaches proposed for white-box and black-box attacks, optimization-based methods~\citep{carlini2017towards,chen2017ead, chen2017zoo,ilyas2018black} are most effective: they usually achieve relatively low distortions and high attack success rates. 
% However, these methods are far from efficient. In the white-box setting, they need to solve constrained optimization problems~\citep{carlini2017towards}, and are usually significantly slower than Fast Gradient Sign Method (FGSM)~\citep{goodfellow6572explaining} or Iterative FGSM (I-FGM)~\citep{kurakin2016adversarial}. 
% Applying those methods with one or two examples are fine, yet in the case of attacking hundreds of thousands examples, e.g., in adversarial training \citep{kurakin2016adversarial, madry2017towards}, this is far from satisfactory.

%The lack of efficiency in white-box attacks also leads to even more severe situations in the black-box setting, 
In the black-box attack, since one needs to make gradient estimations in such setting, a large number of queries are required to perform a successful black-box attack, especially when the data dimension is high. A naive way to estimate gradient direction is to perform finite difference approximation on each dimension~\citep{chen2017zoo}. This would take $O(d)$ queries to performance one full gradient estimation where $d$ is the data dimension and therefore result in inefficient attacks.
For example, attacking a $299\times 299\times 3$ ImageNet \citep{deng2009imagenet} image may take  hundreds of thousands of queries. 
This significantly limits the practical usefulness of such algorithms since they can be easily defeated by limiting the number of queries that an adversary can make to the target model.
Although recent studies \citep{ilyas2018black,ilyas2018prior} have improved the query complexity by using Gaussian sensing vectors or gradient priors, due to the inefficiencies of PGD framework, there is still room for improvements.

%Among the methods proposed for black-box attack, optimization-based approaches~\citep{chen2017zoo,ilyas2018black,cheng2018query} are the most effective: they usually achieve relatively low distortions and high attack success rates. However, these methods are not query efficient since they need to make coordinate-wise gradient estimations. This means a large number of queries are needed for them to perform a successful attack, especially when the data dimension is large. Therefore, they may not be suitable for robustness evaluation since their high query complexities may give us a false sense of target model robustness. 

In this paper, we propose efficient and effective optimization-based adversarial attack algorithms based on a variant of Frank-Wolfe algorithm.
We show in theory that the proposed attack algorithms are efficient with guaranteed convergence rate. The empirical results also verify the efficiency and effectiveness of our proposed algorithms.
% we aim to examine the following questions:
% \textit{Can we improve the efficiency of the optimization-based attack algorithms? In other words, can we use less time and queries to conduct adversarial attacks?} 
% 
% In this work, we answer this question affirmatively. The core idea is to adopt a variant of Frank-Wolfe algorithm to efficiently and effectively solve this attack problem. 
% As another popular constraint optimization tool aside from PGD, Frank-Wolfe algorithm works by calling the linear Minimization Oracle (LMO) over the constraint set at each iteration.

In summary, we make the following main contributions:
%Specifically, our contributions in this work can be summarized as follows:

\begin{enumerate}%[leftmargin=*]
\item We develop a new Frank-Wolfe based projection-free attack framework with momentum mechanism. The framework contains an iterative first-order white-box attack algorithm which admits the fast gradient sign method (FGSM) as a one-step special case, and also a corresponding black-box attack algorithm which adopts zeroth-order optimization with two sensing vector options (either from the Euclidean unit sphere or from the standard Gaussian distribution). 
\item We prove that the proposed white-box and black-box attack algorithms with momentum mechanism enjoy an $O(1/\sqrt{T})$ convergence rate in the nonconvex setting. We also show that the query complexity of the proposed black-box attack algorithm is linear in data dimension $d$.
\item Our experiments on MNIST and ImageNet datasets show that (i) the proposed white-box attack algorithm has better distortion and is more efficient than all the state-of-the-art white-box attack baseline algorithms, and (ii) the proposed black-box attack algorithm is highly query efficient and achieves the highest attack success rate among other baselines.
\end{enumerate}

The remainder of this paper is organized as follows: in Section \ref{sec:related}, we briefly review existing literature on adversarial examples and Frank-Wolfe algorithm. We present our proposed Frank-Wolfe framework in Section \ref{sec:methods}, and the main theory in Section \ref{sec:theory}.  In Section \ref{sec:exp}, we compare the proposed algorithms with state-of-the-art adversarial attack algorithms on ImageNet and MNIST datasets. Finally, we conclude this paper in Section \ref{sec:conclusion}.

%\paragraph{Notation:} 

\section{Related Work}\label{sec:related}
There is a large body of work on adversarial attacks. In this section, we review the most relevant work in both white-box and black-box attack settings, as well as the non-convex Frank-Wolfe optimization.

\noindent\textbf{White-box Attacks:}
\cite{szegedy2013intriguing} proposed to use box-constrained L-BFGS algorithm for conducting white-box attacks. \cite{goodfellow6572explaining} proposed the Fast Gradient Sign Method (FGSM) based on linearization of the network as a simple alternative to L-BFGS. \cite{kurakin2016adversarial} proposed to iteratively perform one-step FGSM \citep{goodfellow6572explaining} algorithm and clips the adversarial point back to the distortion limit after every iteration. It is called Basic Iterative Method (BIM) or I-FGM in the literature. \cite{madry2017towards} showed that for the $L_\infty$ norm case, BIM/I-FGM is almost\footnote{Standard PGD in the optimization literature uses the exact gradient to perform the update step while PGD \citep{madry2017towards} is actually the steepest descent \cite{boyd2004convex} with respect to $L_\infty$ norm. 
} equivalent to Projected Gradient Descent (PGD), which is a standard tool for constrained optimization. \cite{papernot2016limitations} proposed JSMA to greedily attack the most significant pixel based on the  Jacobian-based saliency map. \cite{moosavi2016deepfool} proposed attack methods by projecting the data to the closest separating hyperplane.
\cite{carlini2017towards} introduced the so-called CW attack by proposing multiple new loss functions for generating adversarial examples.  \cite{chen2017ead} followed CW's framework and use an Elastic Net term as the distortion penalty. \cite{dong2018boosting} proposed MI-FGSM to boost the attack performances using momentum.

\noindent\textbf{Black-box Attacks:}
One popular family of black-box attacks~\citep{hu2017generating,papernot2016transferability,papernot2017practical} is based on the transferability of adversarial examples~\citep{liu2016delving,bhagoji2017exploring}, where an adversarial example generated for one DNN may be reused to attack other neural networks. This allows the adversary to construct a substitute model that mimics the targeted DNN, and then attack the constructed substitute model using white-box attack methods.
However, this type of attack algorithms usually suffer from large distortions and relatively low success rates~\citep{chen2017zoo}. To address this issue, \cite{chen2017zoo} proposed the Zeroth-Order Optimization (ZOO) algorithm that extends the CW attack to the black-box setting and uses a zeroth-order optimization approach to conduct the attack. Although ZOO achieves much higher attack success rates than the substitute model-based black-box attacks, it suffers from a poor query complexity since its naive implementation requires to estimate the gradients of all the coordinates (pixels) of the image. To improve its query complexity, several approaches have been proposed. For example, \cite{DBLP:journals/corr/abs-1805-11770} introduces an adaptive random gradient estimation algorithm and 
a well-trained Autoencoder to speed up the attack process. \cite{ilyas2018black} and \cite{liu2016delving} improved ZOO's query complexity by using Natural Evolutionary Strategies (NES) \citep{wierstra2014natural,salimans2017evolution} and active learning, respectively. \cite{ilyas2018prior} further improve the performance by considering the gradient priors.

\noindent\textbf{Non-convex Frank-Wolfe Algorithms:} The Frank-Wolfe algorithm \citep{frank1956algorithm}, also known as the conditional gradient method, is an iterative optimization method for constrained optimization problem. \cite{jaggi2013revisiting} revisited Frank-Wolfe algorithm in 2013 and provided a stronger and more general convergence analysis in the convex setting. \cite{yu2017generalized} proved the first convergence rate for Frank-Wolfe type algorithm in the non-convex setting. \cite{lacoste2016convergence} provided the convergence guarantee for Frank-Wolfe algorithm in the non-convex setting with adaptive step sizes.  \cite{reddi2016stochastic} further studied the convergence rate of non-convex stochastic Frank-Wolfe algorithm in the finite-sum optimization setting. Very recently, \cite{staibdistributionally} proposed to use Frank-Wolfe for distributionally robust training \citep{sinha2018certifying}. \cite{balasubramanian2018zeroth} proved the convergence rate for zeroth-order nonconvex Frank-Wolfe algorithm using one-side finite difference gradient estimator with standard Gaussian sensing vectors.
% \CC{cite the NIPS paper}

\section{Methodology}\label{sec:methods}
\subsection{Notation}
Throughout the paper, scalars are denoted by lower case letters,  vectors by lower case bold face letters and sets by calligraphy upper cae letters. For a vector $\xb \in \RR^d$, we denote the $L_p$ norm of $\xb$ by $\| \xb \|_p = (\sum_{i=1}^d x_i^p)^{1/p}$. Specially, for $p = \infty$, the $L_\infty$ norm of $\xb$ by $ \|\xb\|_\infty = \max_{i=1}^d |\theta_i|$. We denote $\cP_\cX(\xb)$ as the projection operation of projecting vector $\xb$ into the set $\cX$.

\subsection{Problem Formulation}
According to the attack purposes, attacks can be divided into two categories: \textit{untargeted attack} and \textit{targeted attack}. 

In particular, untargeted attack aims to turn the prediction into any incorrect label, while the targeted attack, requires to mislead the classifier to a specific target class. 
% Our proposed algorithms can work for both untargeted attack and targeted attack straightforwardly.
In this work, we focus on the strictly harder targeted attack setting \citep{carlini2017towards,ilyas2018black}.
It is worth noting that our proposed algorithm can be extended to untargeted attack straightforwardly.
To be more specific, let us define $\ell(\xb, y)$ as the classification loss function of the targeted DNN with an input $\xb \in \RR^d$ and a corresponding label $y$. 
For targeted attacks, we aim to minimize $\ell(\xb, y_{\text{tar}})$ to learn an adversarial example that will be misclassified to the target class $y_{\text{tar}}$. %And for untargeted attacks, we aim to minimize $-\ell(\xb, y_{\text{true}})$ to learn an adversarial example that is different from the true class label $y_{\text{true}}$. To unify both cases, 
In the rest of this paper, let $f(\xb)=\ell(\xb, y_{\text{tar}})$ be the attack loss function for simplicity, and the corresponding targeted attack problem\footnote{Note that there is usually an additional constraint on the input variable $\xb$, e.g., $\xb \in [0,1]^n$ for normalized image inputs.} can be formulated as the following optimization problem:
\begin{eqnarray}\label{eq:opt_problem}
    &\min_{\xb} & f(\xb) \notag\\
    &\mbox{subject to}& \|\xb - \xb_{\ori} \|_p \leq \epsilon.
\end{eqnarray}
Evidently, the constraint set $\cX := \{\xb \ | \  \|\xb - \xb_{\ori}\|_p \leq \epsilon\}$ is a bounded convex set when $p\ge 1$. 
Note that even though we mainly focus on the most popular $L_\infty$ attack case in this paper, our proposed methods can easily extend to general $p\ge 1$ case.

\subsection{Frank-Wolfe vs. PGD}
% This actually deviates from the original idea of FGSM, linearizing the network. That makes us wonder what if we follow the idea of FGSM and perform linearization in each step iteratively. This actually coincides with the idea of Frank-Wolfe algorithm \citep{frank1956algorithm} (also known as the conditional gradient descent), which is another popular optimization tool for constrained optimization.
Although PGD can achieve relatively high attack success rate within moderate iterates, the multi-step update formula requires an additional projection step at each iteration to keep the iterates within the constraint set. This tends to cause the generated adversarial examples near or upon the boundary of the constraint set, and leads to relatively large distortion.
This motivates us to use Frank-Wolfe based optimization algorithm \cite{frank1956algorithm}. Different from PGD, Frank-Wolfe algorithm is projection-free as it calls a Linear Minimization Oracle (LMO) over the constraint set $\cX$ at each iteration, i.e., 
\begin{align*}
    {\rm{LMO}} \in \argmin_{\xb \in \cX} \langle \xb, \nabla f(\xb_t) \rangle.
\end{align*}
The LMO can be seen as the minimization of the first-order Taylor expansion of $f(\cdot)$ at point $\xb_t$:
\begin{align*}
    \min_{\xb \in \cX} f(\xb_t) + \langle \xb - \xb_t, \nabla f(\xb_t) \rangle.
\end{align*}
By calling LMO, Frank Wolfe solves the linear problem in $\cX$ and then perform weighted average with previous iterate to obtain the final update formula. 

Comparing the two methods, PGD is a more ``aggressive'' approach. It first takes a step towards the negative gradient direction while ignoring the constraint to get a new point (often outside the constraint set), and then correct the new point by projecting it back into the constraint set. 
In sharp contrast, Frank-Wolfe is more ``conservative'' as it always keeps the iterates within the constraint set. Therefore, it avoids projection and can lead to better distortion. 
% One another thing is PGD can only deal with simple $L_\infty$ or $L_2$ norm constraint case while for general $p$-norm constraint, the projection step can be hard to solve.

\subsection{Frank-Wolfe White-box Attacks}
The proposed Frank-Wolfe based white-box attack algorithm is shown in Algorithm \ref{alg:fw}, which is built upon the classic Frank-Wolfe algorithm. 
The key difference between Algorithm \ref{alg:fw} and the classic Frank-Wolfe algorithm is in Line \ref{op:momentum}, where an additional momentum term $\mb_t$ is introduced.
% As pointed out by \cite{lacoste2015global}, when the optimal solution lies near the boundary, classic Frank-Wolfe algorithm start to zig-zag, causing the convergence to slow down.
% The motivation here is to stabilize the LMO direction by replacing the exact gradient with an exponentially averaged gradient. 
The momentum term $\mb_t$ will help stabilize the LMO direction and leads to empirically accelerated convergence of Algorithm \ref{alg:fw}.

\begin{algorithm}[ht]
	\caption{Frank-Wolfe White-box Attack Algorithm}
	\label{alg:fw}
	\begin{algorithmic}[1]
		\STATE \textbf{input:} number of iterations $T$, step sizes $\{\gamma_t\}$;
        \STATE $ \xb_0 = \xb_{\ori}, \mb_{-1} = \nabla f(\xb_0) $
 		\FOR {$t = 0,\ldots, T-1$}
 		     \STATE $\mb_t = \beta \cdot \mb_{t-1} + (1 - \beta) \cdot \nabla f(\xb_t)$ \label{op:momentum}
		     \STATE $\vb_t = \argmin_{\xb \in \cX} \langle \xb, \mb_t \rangle $ \  // LMO \label{op:lmo}
		     \STATE $ \db_t = \vb_t - \xb_t $
		     \STATE $ \xb_{t+1} = \xb_t + \gamma_t \db_t $
		\ENDFOR    
		\STATE \textbf{output: } $\xb_T$
	\end{algorithmic}
\end{algorithm}

The LMO solution itself can be expensive to obtain in general. Fortunately, %applying Frank-Wolfe to solve \eqref{eq:opt_problem} actually gives us 
for the constraint set $\cX$ defined in \eqref{eq:opt_problem}, the corresponding LMO has a closed-form solution. 
Here we provide the closed-form solution of LMO (Line \ref{op:lmo} in Algorithm \ref{alg:fw}) for $L_\infty$ norm case\footnote{The derivation can be found in the Appendix \ref{sec:lmo_dev}.}:
\begin{align*}
% \tag{$L_2$ norm}
    % &\vb_t = -\frac{\epsilon \cdot \mb_t}{\|\mb_t\|_2}  + \xb_{\ori}, \\
    % \tag{$L_\infty$ norm}
    &\vb_t =  - \epsilon \cdot \sign( \mb_t) + \xb_{\ori}.
\end{align*}
Note that if we write down the full update formula at each iteration in Algorithm \ref{alg:fw}, it becomes
\begin{align}\label{eq:full_update}
    \xb_{t+1} = \xb_t - \gamma_t \epsilon \cdot \sign(\mb_t) - \gamma_t(\xb_t - \xb_{\ori}).
\end{align}
Intuitively speaking, the term $-\gamma_t(\xb_t - \xb_{\ori})$ enforces $\xb_t$ to be close to $\xb_{\ori}$ for all $t=1,\ldots,T$, which encourages the adversarial example to have a small distortion. This is the key advantage of  Algorithm \ref{alg:fw}.
 
% and for $p =2$ case, it takes
% \begin{align*}
%     \xb_{t+1}  
%     &= \xb_t - \gamma_t \epsilon \cdot \frac{ \nabla f(\xb_t)}{\|\nabla f(\xb_t)\|_2} - \gamma_t(\xb_t - \xb_{\ori}).
% \end{align*}

\noindent\textbf{Comparison with FGSM:} When $T =1$, substituting the above LMO solutions into Algorithm \ref{alg:fw} yields the final update of $\xb_1 = \xb_0 - \gamma_t \epsilon \cdot \sign(\nabla f(\xb_0))$, which reduces to FGSM\footnote{The extra clipping operation in FGSM is to project to the additional box constraint for image classification task. We will also need this clipping operation at the end of each iteration for specific tasks such as image classification.} when $\gamma_t = 1$. 
Therefore, our proposed Frank-Wolfe white-box attack also includes FGSM as a one-step special instance.

\subsection{Frank-Wolfe Black-box Attacks}
Next we consider the black-box setting, where we cannot perform back-propagation to calculate the gradient of the loss function anymore. Instead, we can only query the DNN system's outputs with specific inputs. To clarify, here the output refers to the logit layer's output (confidence scores for classification), not the final prediction label. 
% The label-only setting is doable under our framework, but will incur extra difficulty such as designing new loss functions. For simplicity, here we consider the confidence score output.

We propose a zeroth-order Frank-Wolfe based algorithm to solve this problem in Algorithm \ref{alg:fw-blackbox}. %shows our proposed Frank-Wolfe black-box attack algorithm. 
The key difference between our proposed black-box attack and white-box attack is one extra gradient estimation step, which is presented in Line \ref{op:grad_est} in Algorithm \ref{alg:fw-blackbox}. 
Also, the momentum term $\mb_t$ is now defined as the exponential average of previous gradient estimations $\{\qb_t\}_{t=0}^{T-1}$. This will help reduce the variance in zeroth-order gradient estimation and empirically accelerate the convergence of Algorithm \ref{alg:fw-blackbox}. 
% Note that for the final output, we provide two options. While option II is the common choice in practice, option I is also provided for the ease of theoretical analysis. 

\begin{algorithm}[ht]
	\caption{Frank-Wolfe Black-box Attack Algorithm}
	\label{alg:fw-blackbox}
	\begin{algorithmic}[1]
		\STATE \textbf{input:} number of iterations $T$, step sizes $\{\gamma_t\}$, sample size for gradient estimation $b$, sampling parameter $\delta$;
        \STATE $ \xb_0 = \xb_{\ori} $, $\mb_{-1} = {\rm GRAD\_EST} (\xb_0,b,\delta)$
 		\FOR {$t = 0,\ldots, T-1$}
 		     \STATE $\qb_t =  {\rm GRAD\_EST} (\xb_t,b,\delta)$ \  // Alg \ref{alg:grad_est} \label{op:grad_est}
 		     \STATE $\mb_t = \beta \cdot \mb_{t-1} + (1 - \beta) \cdot \qb_t$ 
		     \STATE $\vb_t = \argmin_{\vb \in \cX} \langle \vb, \mb_t \rangle $
		     \STATE $ \db_t = \vb_t - \xb_t $
		     \STATE $ \xb_{t+1} = \xb_t + \gamma_t \db_t $
		\ENDFOR    
% 		\STATE \textbf{option I:} $\xb_a $ is uniformly random chosen from $\{\xb_t\}_{t=1}^T$
% 	    \STATE \textbf{option II:} $\xb_a = \xb_T$
% 		\STATE \textbf{output: } $\xb_a$
		\STATE \textbf{output: } $\xb_T$
	\end{algorithmic}
\end{algorithm}

As in many other zeroth-order optimization algorithms \citep{shamir2017optimal,flaxman2005online}, Algorithm \ref{alg:grad_est} uses symmetric finite differences to estimate the gradient and therefore, gets rid of the dependence on back-propagation in white-box setting. Different from \cite{chen2017zoo}, here we do not utilize natural basis as our sensing vectors, instead, we provide two options: one is to use vectors uniformly sampled from Euclidean unit sphere and the other is to use vectors uniformly sampled from standard multivarite Gaussian distribution. This will greatly improve the gradient estimation efficiency comparing to sensing with natural basis as such option will only be able to estimate one coordinate of the gradient vector per query.
In practice, both options here provide us competitive experimental results.
It is worth noting that NES method \citep{wierstra2014natural} with antithetic sampling \citep{salimans2017evolution} used in \cite{ilyas2018black} yields similar formula as our option II in Algorithm \ref{alg:grad_est}.

\begin{algorithm}[ht]
	\caption{{\rm GRAD\_EST}($\xb,b,\delta$)}
	\label{alg:grad_est}
	\begin{algorithmic}[1]
		%\STATE \textbf{parameters:} 
 		     \STATE $\qb = \zero$
 		     \FOR {$i = 1,\ldots, b$}
 		        \STATE \textbf{option I:} Sample $\ub_i$ uniformly from the Euclidean unit sphere with $\|\ub_i\|_2 = 1$ \\
 		         \quad \ $\qb = \qb +  \frac{d}{2\delta b} \big(  f(\xb + \delta \ub_i) - f(\xb - \delta \ub_i) \big) \ub_i$
 		        \STATE \textbf{option II:} Sample $\ub_i$ uniformly from the standard Gaussian distribution $N(\zero, \Ib)$ \\
 		        \quad \ $\qb = \qb +  \frac{1}{2\delta b} \big(  f(\xb + \delta \ub_i) - f(\xb - \delta \ub_i) \big) \ub_i$
 		     \ENDFOR
 
		\STATE \textbf{return} $\qb$
	\end{algorithmic}
\end{algorithm}

\section{Main Theory}\label{sec:theory}
In this section, we establish the convergence guarantees for our proposed Frank-Wolfe adversarial attack algorithms described in Section \ref{sec:methods}. The omitted proofs can be found in the Appendix \ref{sec:proof}.
First, we introduce the convergence criterion for our Frank-Wolfe adversarial attack framework.

\subsection{Convergence Criterion}
The loss function for common DNN models are generally nonconvex. In addition, \eqref{eq:opt_problem} is a constrained optimization. For such general nonconvex constrained optimization, we typically adopt the Frank-Wolfe gap as the convergence criterion (since gradient norm of $f$ is no longer a proper criterion for constrained optimization problems): 
\begin{align*}
    g(\xb_t) = \max_{\xb \in \cX} \langle \xb - \xb_t, -\nabla f(\xb_t) \rangle. 
\end{align*}
Note that we always have $g(\xb_t) \geq 0$ and $\xb_t$ is a stationary point for the constrained optimization problem if and only if $g(\xb_t) = 0$, which makes $g(\xb_t)$ a perfect convergence criterion for Frank-Wolfe based algorithms.

\subsection{Convergence Guarantee for Frank-Wolfe White-box Attack}
%Now .
Before we are going to provide the convergence guarantee of Frank-Wolfe white-box attack (Algorithm \ref{alg:fw}), we introduce the following assumptions that are essential to the convergence analysis.
\begin{assumption}
\label{assump:smooth}
Function $f(\cdot)$ is $L$-smooth with respect to $\xb$, i.e., for any $\xb, \xb'$, it holds that
\begin{align*}
    f(\xb') \leq f(\xb) + \nabla f(\xb)^\top (\xb' - \xb) + \frac{L}{2}\|\xb' -\xb\|_2^2. 
\end{align*}
\end{assumption}

Assumption \ref{assump:smooth} is a standard assumption in nonconvex optimization, and is also adopted in other Frank-Wolfe literature such as \cite{lacoste2016convergence,reddi2016stochastic}.
Note that even though the smoothness assumption does not hold for general DNN models, a recent study \citep{santurkar2018does} shows that batch normalization that is used in many modern DNNs such as Inception V3 model, actually makes the optimization landscape significantly smoother\footnote{The original argument in \cite{santurkar2018does} refers to the smoothness with respect to each layer's parameters. Note that the first layer's parameters are in the mirror position (in terms of backpropagation) as the network inputs. Therefore, the argument in \cite{santurkar2018does}  can also be applied here with respect to the network inputs.}. This justifies the validity of Assumption \ref{assump:smooth}.

%\CC{Since smoothness assumption does not hold in most neural networks, you need to emphasize that this assumption is mainly for the purpose of simplified analysis. Also, check out this paper: How Does Batch Normalization Help Optimization? (No, It Is Not About Internal Covariate Shift). You can cite it and make an argument that batch normalization make the loss function smooth. Of course, the smoothness is w.r.t. parameter, not to data. So it's tricky}

\begin{assumption}
\label{assump:bounded_domain}
Set $\cX$ is bounded with diameter $D$, i.e., $\|\xb -\xb'\|_2 \leq D$ for all $\xb, \xb' \in \cX$.
\end{assumption}

Assumption \ref{assump:bounded_domain} implies that the input space is bounded. For common tasks such as image classification, given the fact that images have bounded pixel range and $\epsilon$ is a small constant, this assumption trivially holds. 
Given the above assumptions, the following lemma shows that the momentum term $\mb_t$ will not deviate from the gradient direction significantly.
\begin{lemma}\label{lemma:m_t}
Under Assumptions \ref{assump:smooth} and \ref{assump:bounded_domain}, for $\mb_t$ in Algorithm \ref{alg:fw}, it holds that
\begin{align*}
     \|\nabla f(\xb_t) - \mb_t\|_2 \leq \frac{\gamma LD}{1 - \beta}.
\end{align*}
\end{lemma}

Now we present the theorem, which characterizes the convergence rate of our proposed Frank-Wolfe white-box adversarial attack algorithm presented in Algorithm \ref{alg:fw}.
\begin{theorem}\label{theorem:white}
Under Assumptions \ref{assump:smooth} and \ref{assump:bounded_domain}, let $\gamma_t = \gamma = \sqrt{ 2(f(\xb_0) - f(\xb^*))/( C_\beta LD^2T)}$, the output of Algorithm \ref{alg:fw} satisfies
\begin{align*}
    \tilde g_T  
    &\leq \sqrt{ \frac{2C_\beta LD^2  (f(\xb_0) - f(\xb^*))}{ T}} ,
\end{align*}
where $\tilde g_T = \min_{1 \leq k \leq T} g(\xb_k)$, $\xb^*$ is the optimal solution to \eqref{eq:opt_problem} and $C_\beta = (3-\beta)/(1-\beta)$.
\end{theorem}

\begin{remark}
Theorem \ref{theorem:white} suggests that our proposed Frank-Wolfe white-box attack algorithm achieves a $O(1/\sqrt{T})$ rate of convergence. 
Unlike previous work \citep{lacoste2016convergence} which focuses on the convergence rate of classic Frank-Wolfe method, our analysis shows the convergence rate of the Frank-Wolfe method with momentum mechanism.
\end{remark}

\subsection{Convergence Guarantee for Frank-Wolfe Black-box Attack}
Next we analyze the convergence of our proposed Frank-Wolfe black-box adversarial attack algorithm presented in Algorithm \ref{alg:fw-blackbox}. 

In order to prove the convergence of our proposed Frank-Wolfe black-box attack algorithm, we need the following additional assumption that $\|\nabla f(\zero)\|_2$ is bounded.
\begin{assumption}
\label{assump:grad_zero_bound}
Gradient of $f(\cdot)$ at zero point $\nabla f(\zero)$ satisfies  $\max_y \|\nabla f(\zero)\|_2 \leq G$.
\end{assumption}

% This assumption is also easy to satisfy since ...

Following the analysis in \cite{shamir2017optimal}, let $f_\delta(\xb) = \EE_\ub[f(\xb + \delta \ub) ]$, which is the smoothed version of $f(\xb)$. This smoothed function value plays a central role in our theoretical analysis, since it bridges the finite difference gradient approximation with the actual gradient. The following lemma shows this relationship.

\begin{lemma}
\label{lemma:zero_order}
For any $\xb$ and the gradient estimator $\qb$ of $\nabla f(\xb)$ in Algorithm \ref{alg:grad_est}, its expectation and variance satisfy
\begin{align*}
    &\EE [\qb] = \nabla f_\delta(\xb),\\
    &\EE \|\qb  - \EE[\qb]\|_2^2 \leq  \frac{1}{b}  \bigg( 2d(G+ LD)^2 + \frac{1}{2}\delta^2 L^2 d^2 \bigg).
\end{align*}
And also we have
\begin{align*} 
    \EE \|\nabla f(\xb) - \qb\|_2  
    &\leq \frac{\delta Ld}{2} + \frac{ 2\sqrt{d}(G+ LD) +  \delta L d }{\sqrt{2b}}  .
\end{align*}
\end{lemma}

Now we are going to present the theorem, which characterizes the convergence rate of Algorithm \ref{alg:fw-blackbox}.

\begin{theorem}\label{theorem:black}
Under Assumptions \ref{assump:smooth}, \ref{assump:bounded_domain} and \ref{assump:grad_zero_bound}, let $\gamma_t =\gamma = \sqrt{(f(\xb_0) - f(\xb^*))/( C_\beta LD^2T)}$, $b = Td$ and $\delta = \sqrt{1/(Td^2)}$, the output of Algorithm \ref{alg:fw-blackbox} satisfies
\begin{align*}
    \EE [\tilde g_T]  \leq\frac{D}{ \sqrt{T}} \Big ( 2\sqrt{C_\beta L(f(\xb_0) - f(\xb^*))} + C_\beta (L +G+ LD)  \Big) ,
\end{align*}
where $\tilde g_T = \min_{1 \leq k \leq T} g(\xb_k)$, the expectation of $\tilde g_T$ is over the randomness of the gradient estimator, $\xb^*$ is the optimal solution to \eqref{eq:opt_problem} and $C_\beta = (3-\beta)/(1-\beta)$.
\end{theorem}

\begin{remark}
Theorem \ref{theorem:black} suggests that Algorithm \ref{alg:fw-blackbox} also enjoys a $O(1/\sqrt{T})$ rate of convergence. 
Note that \cite{balasubramanian2018zeroth} proves the convergence rate for classic zeroth-order Frank-Wolfe algorithm. Our result is different in several aspects. First, we prove the convergence rate of zeroth-order Frank-Wolfe with momentum. Second, we use symmetric finite difference gradient estimator with two types of sensing vectors while they \cite{balasubramanian2018zeroth} use one-side finite difference gradient estimator with Gaussian sensing vectors.
In terms of query complexity, the total number of queries needed in Algorithm \ref{alg:fw-blackbox} is $Tb = T^2d$, which is linear in the data dimension $d$. In fact, in the experiment part, we observe that this number can be substantially smaller than $d$, e.g., $b = 25$. %Furthermore, although we only prove the convergence rate of Algorithm \ref{alg:fw-blackbox} with option I, our result can be readily extended to Algorithm \ref{alg:fw-blackbox} with option II (the Gaussian sensing vector case). 
\end{remark}

\section{Experiments}\label{sec:exp}
In this section, we present the experimental results for our proposed Frank-Wolfe attack framework against other state-of-the-art adversarial attack algorithms in both white-box and black-box settings. All of our experiments are conducted on Amazon AWS p3.2xlarge servers which come with Intel Xeon E5 CPU and one NVIDIA Tesla V100 GPU (16G RAM). All experiments are implemented in Tensorflow platform version $1.10.0$ within Python $3.6.4$.

\subsection{Evaluation Setup}
%\cc{
We compare the performance of all attack algorithms by evaluating on both MNIST \citep{lecun1998mnist} and ImageNet \citep{deng2009imagenet} datasets.
For MNIST dataset, we attack a pre-trained 6-layer CNN: 4 convolutional layers followed by 2 dense layers with max-pooling and Relu activations applied after each convolutional layer. The pre-trained model achieves $99.3\%$ accuracy on MNIST test set. 
For ImageNet experiments, we attack a pre-trained Inception V3 model \citep{szegedy2016rethinking}.
The pre-trained Inception V3 model is reported to have a $78.0\%$ top-$1$ accuracy and a $93.9\%$ top-$5$ accuracy. 
% The pre-trained ResNet V2 model is reported to have a $75.6\%$ top-$1$ and a $92.8\%$ top-$5$ accuracy.
For MNIST dataset, we randomly choose $1000$ images from its test set that are verified to be correctly classified by the pre-trained model and also randomly choose a target class for each image.
Similarly, for ImageNet dataset, we randomly choose $250$ images from its validation set 
as our attack examples. 
% that are verified to be correctly classified by the pre-trained model and also randomly choose a target class for each image. 
% In terms of the constraint set, we set $\epsilon = 0.05$ for ImageNet dataset and $\epsilon = 0.3$ for MNIST dataset.
%}
For our proposed black-box attack, we test both options in Algorithm \ref{alg:grad_est}. We performed grid search to tune the hyper-parameters for all algorithm to ensure a fair comparison.
Detailed description on hyperparameter tuning and parameter settings can be found in the Appendix \ref{sec:exp_setting}.

\subsection{Baseline Methods}
%\cc{
We compare the proposed algorithms with several state-of-the-art baseline algorithms. Specifically, we compare the proposed white-box attack algorithm 
with 
% \footnote{We did not compare with FGM (FGSM) \citep{goodfellow6572explaining} since it basically has zero success rate for targeted attack on Inception V3 model or ResNet V2 model. }
(i) FGSM \citep{goodfellow6572explaining}
(ii) PGD \citep{madry2017towards} (normalized steepest descent\footnote{
% It is not exactly standard PGD in optimization field. PGD \citep{madry2017towards} uses signed (normalized) gradient which has balanced scale in different part of the constraint set.
standard PGD will need large step size to go anywhere since the gradient around the true example is relatively small. On the other hand, the large step size will cause the algorithm go out of the constraint set quickly and basically stop moving since then because of the projection step.
})
% (ii) CW attack \citep{carlini2017towards} and (iii) EAD attack \citep{chen2017ead}. 
(iii) MI-FGSM \citep{dong2018boosting}.
We compare the proposed black-box attack algorithm with
% (i) ZOO attack \citep{chen2017zoo},
(i) NES-PGD attack \citep{ilyas2018black} and (ii) Bandit attack \citep{ilyas2018prior}. We did not report the comparison with ZOO \citep{chen2017zoo} here because it consistently underperforms NES-PGD and Bandit attacks according to our experiments and prior work.
% We did not compare with \cite{ilyas2018prior} since that it improves the NES-PGD attack by considering the gradient priors, which is orthogonal to our focus in this paper, and similar idea can also be incorporated into our Frank-Wolfe framework. We leave it as a future work.
%}

\subsection{White-box Attack Experiments}
In this subsection, we present the white-box attack experiments on both MNIST and ImageNet datasets. We choose $\epsilon = 0.3$ for MNIST dataset and $\epsilon = 0.05$ for ImageNet dataset. 
For comparison, we 
% limit the maximum number of gradient iterations in attacking each sample to $1000$ and 
report the attack success rate, average number of iterations to complete the attack, as well as average distortion for each method.

%\cc{ 
Tables \ref{table:white_linf_mnist} and \ref{table:white_linf_imagenet} present our experimental results for the white-box attack experiments. 
For experiments on both datasets, while FGSM only needs $1$ gradient update per attack, it only achieves $21.5\%$ attack success rate on MNIST and $1.2\%$ attack success rate on ImageNet in the targeted attack setting. All the other methods achieve $100\%$ attack success rate. PGD needs in average $6.2$ and $8.7$ gradient iterations per attack on MNIST and ImageNet respectively. MI-FGSM improves it to around $4.0$ and $5.0$ iterations per attack on MNIST and ImageNet. However, the distortion of both PGD and MI-FGSM is very close to the perturbation limit $\epsilon$, which indicates that their generated adversarial examples are near or upon the boundary of the constraint set.
On the other hand, our proposed Frank-Wolfe white-box attack algorithm achieves not only the smallest average number of iterations per attack, but also the smallest distortion among the baselines. This suggests the advantage of Frank-Wolfe based projection-free algorithms for white-box attack.
% also achieves over two times smaller distortion compared with the baselines. 
% For MNIST experiment, similar patterns can be observed. FGSM achieves $21.5\%$ attack success rate while all other methods achieve $100\%$ attack success rate. Our proposed Frank-Wolfe white-box attack algorithm, again, achieves the smallest average iterations per attack and the smallest distortion among the baselines. 
% CW method takes significantly longer time and does not perform very well on average distortion either. This is largely due to the original CW was designed for $L_2$ norm attack, and in order to apply it to $L_\infty$ norm attack, special design is needed, which sacrifices its performance in terms of runtime.
%}

% \vspace{-0.4cm}
\begin{table}[h!]
  \caption{Comparison of targeted $L_\infty$ norm based white-box attacks on MNIST dataset with $\epsilon = 0.3$.
  }
  \label{table:white_linf_mnist}
  \begin{center}
  \begin{small}
  \begin{tabular}{l|ccc}
    \toprule
    Methods & ASR(\%) & \# Iterations & Distortion\\
    \midrule
    FGSM    & 21.5  & - & 0.300\\
    PGD     & 100.0 & 6.2 & 0.277 \\
    MI-FGSM & 100.0 & 4.0 & 0.279\\
    % CW  & 100.0 &  -  & -      & 100.0 & -   &  -  \\
    FW-white  &  100.0 &  \textbf{3.3} & \textbf{0.256}  \\
    \bottomrule
  \end{tabular}
  \end{small}
  \end{center}
\end{table}

% \vspace{-0.8cm}
\begin{table}[h!]
  \caption{Comparison of targeted $L_\infty$ norm based white-box attacks on ImageNet dataset with $\epsilon = 0.05$.
  }
  \label{table:white_linf_imagenet}
  \begin{center}
  \begin{small}
  \begin{tabular}{l|ccc}
    \toprule
    Methods & ASR(\%) & \# Iterations & Distortion\\
    \midrule
    FGSM & 1.2 & - & 0.050\\
    PGD & 100.0 & 8.7 &  0.049 \\
    MI-FGSM & 100.0 & 5.0 & 0.049 \\
    % CW  & 100.0 &  -  & -   \\
    FW-white  & 100.0 & \textbf{4.8} & \textbf{0.019} \\
    \bottomrule
  \end{tabular}
  \end{small}
  \end{center}
\end{table}

\begin{table*}[h!]
\caption{Comparison of targeted $L_\infty$ norm based black-box attacks on MNIST and ImageNet datasets in terms of attack success rate, average time and average number of queries (QUERIES: for all images including both successfully and unsuccessfully attacked ones; QUERIES(SUCC): for successfully attacked ones only) needed per image.
}
\label{table:black_linf}
\begin{center}
\begin{scriptsize}
\begin{sc}
\begin{tabular}{l|cccc|cccc}
    \toprule
    \multicolumn{1}{c}{\multirow{2}{*}{Methods}} & \multicolumn{4}{c}{MNIST ($\epsilon = 0.3$)} &
    \multicolumn{4}{c}{ImageNet ($\epsilon = 0.05$)}\\
    % \cline{2-5}
    \cmidrule(r){2-5}
    \cmidrule(r){6-9}
    \multicolumn{1}{c}{} & ASR(\%) & Time(s) & Queries & Queries(succ) & ASR(\%) & Time(s) & Queries & Queries(succ) \\
    
% \begin{tabular}{lcccc}
% \toprule
% Methods & ASR & Time(s) & Queries & Queries(succ)\\
\midrule 
NES-PGD  & 96.8 & 0.2 & 5349.0 & 3871.3 & 88.0 & 85.1 & 26302.8 & 23064.5\\
Bandit  & 86.1 & 4.8 & 8688.9 & 2019.7 & 72.0 & 148.7 & 27172.5 & 18295.2 \\
FW (Sphere)  & \textbf{99.9} & \textbf{0.1} & \textbf{1132.6} & \textbf{1083.6} & 97.2 & 62.1 & 15424.0 & \textbf{14430.8}\\
FW (Gaussian) & \textbf{99.9}  & \textbf{0.1} & 1144.4 & 1095.4 & \textbf{98.4} & \textbf{50.6} &   \textbf{15099.4 } & 14532.3 \\
\bottomrule
\end{tabular}
\end{sc}
\end{scriptsize}
\end{center}
\vskip -0.2in
\end{table*}

\subsection{Black-box Attack Experiments}
% In this subsection, we present the black-box attack experiments on Inception V3 model and ResNet V2 model. The maximum query limit is set to be $50,000$ per attack. 
% Under the query budget, we skip the grid search / binary search steps that are used in the white-box setting since extra queries are needed for finding parameters that can obtain better distortions. For $L_2$ norm based attack, we set $\epsilon = 5$ and for $L_\infty$ based attack, we set $\epsilon = 0.05$.

In this subsection, we present the black-box attack experiments on both MNIST and ImageNet datasets. The maximum query limit is set to be $50,000$ per attack. We choose $\epsilon = 0.3$ for MNIST dataset and $\epsilon = 0.05$ for ImageNet dataset. For comparison, we report the attack success rate, average attack time, average number of queries needed, as well as average number of queries needed on successfully attacked samples for each method.
 
Table \ref{table:black_linf} presents our experimental results for targeted black-box attacks on both ImageNet and MNIST datasets.
% We can observe that ZOO is quite slow in this task. Attack on a single image can take up to $8$ times more time comparing with others and it only achieve a success rate less than  $10.0\%$ under the query budget\footnote{the original paper reports high success rate with a query budget of $10^6$ per attack, which is $20$ times larger than our budget.}. 
We can see that on MNIST, 
NES-PGD method achieves a relatively high attack success rate, but still takes quite a lot queries per (successful) attack. Bandit method improves the query complexity for successfully attacked samples but has lower attack success rate in this setting and takes longer time to complete the attack. 
In sharp contrast, our proposed Frank-Wolfe black-box attack algorithms (both sphere and Gaussian sensing vector options) achieve the highest success rate in the targeted black-box attack setting 
while greatly improve the query complexity by around $50\%$ over the best baseline.
On ImageNet, similar patterns can be observed: our proposed Frank-Wolfe black-box attack algorithms achieve the highest attack success rate and further significantly improve the query efficiency against the baselines. This suggests the advantage of Frank-Wolfe based projection-free algorithms for black-box attack.

\begin{figure}[h] 
\centering
\subfigure[MNIST]{\includegraphics[width=0.4\textwidth]{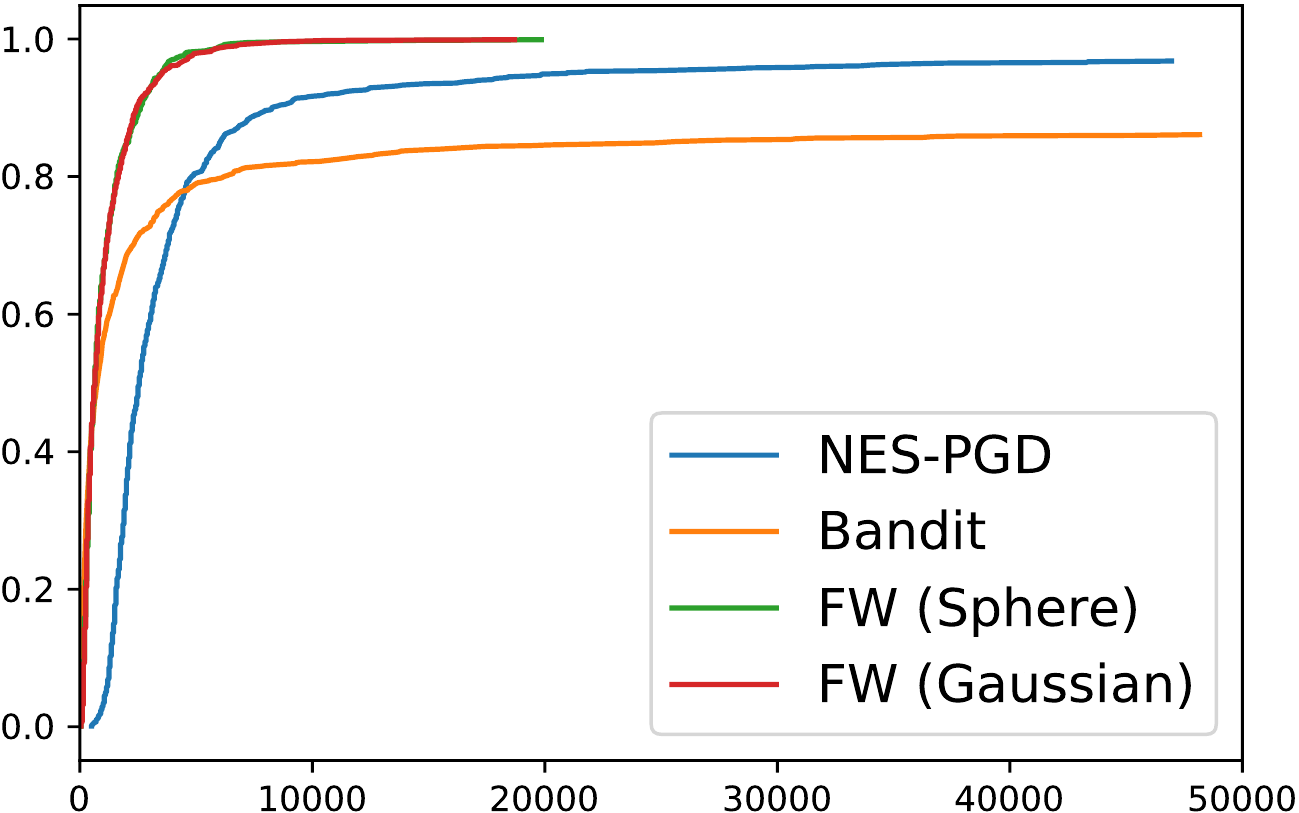}}
\subfigure[ImageNet]{\includegraphics[width=0.4\textwidth]{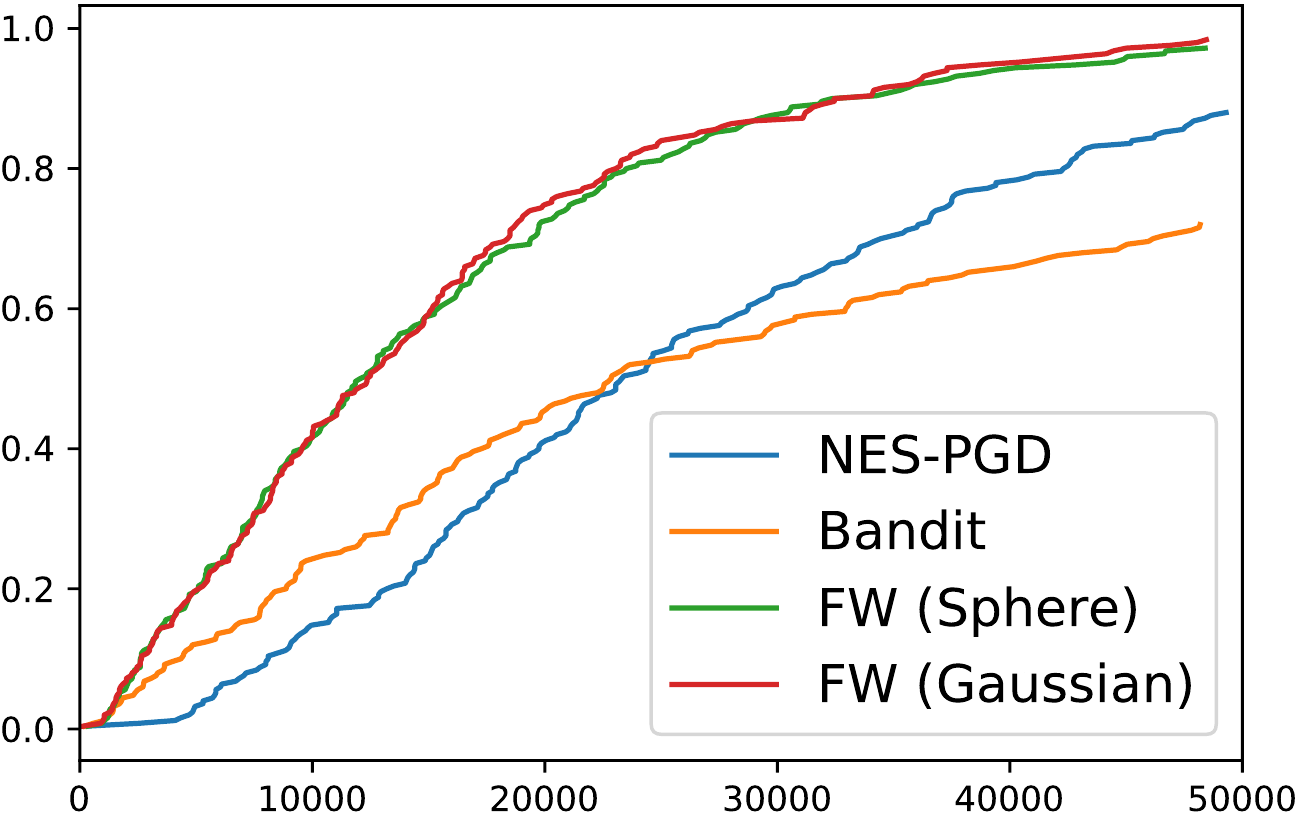}}
\caption{Attack success rate against the number of queries plot for targeted black-box attacks on MNIST and ImageNet datasets.}
\label{fig:black_query_plot}
\end{figure}

To provide more intuitive demonstrations, we also plot the attack success rate against the number of queries for our black-box experiments.
Figure \ref{fig:black_query_plot} shows the plots of the attack success rate against the number of queries for different algorithms on MNIST and ImageNet datasets respectively. As we can see from the plots, 
Bandit attack achieves better query efficiency for easy-to-attack examples (require less queries to attack) compared with NES-PGD or even FW at the early stages, but falls behind even to NES-PGD on hard-to-attack examples (require more queries to attack).
We conjecture that in targeted attack setting, the gradient/data priors are not as accurate as in untargeted attack case, which makes Bandit attack less effective especially on hard-to-attack examples. 
On the other hand, our proposed Frank-Wolfe black-box attack algorithms achieve the highest attack success rate and the best efficiency (least queries needed for achieving the same success rate).  This again confirm the advantage of Frank-Wolfe based projection-free algorithms for black-box attack.
 
% \vspace{-0.2cm}
% \begin{figure}[h!] 
% \centering
% {\includegraphics[width=0.3\textwidth]{mnist_linf_query.pdf}}
% \caption{Attack success rate against the number of queries plot for targeted black-box attacks on MNIST dataset.}
% \label{fig:black_mnist_query_plot}
% \end{figure}

% \vspace{-0.5cm}
% \begin{figure}[h!] 
% \centering
% {\includegraphics[width=0.3\textwidth]{inception_linf_query.pdf}}
% \caption{Attack success rate against the number of queries plot for targeted black-box attacks on ImageNet dataset.}
% \label{fig:black_inception_query_plot}
% \end{figure}

% \begin{figure} 
% \centering
% \subfigure[$L_2$ norm based attack ]{\includegraphics[width=0.23\textwidth]{resnet_l2_query.pdf}}
% \subfigure[$L_\infty$ norm based attack]{\includegraphics[width=0.23\textwidth]{resnet_linf_query.pdf}}
% \caption{Attack success rate against the number of queries plot for different algorithms in both $L_2$ norm and $L_\infty$ norm based black-box attacks on ResNet V2 model.}
% \label{fig:black_resnet_query_plot}
% \end{figure}

% \vspace{-0.2cm}
\section{Conclusions and Future Work}\label{sec:conclusion}
In this work, we propose a Frank-Wolfe framework for efficient and effective  adversarial attacks. Our proposed white-box and black-box attack algorithms enjoy an $O(1/\sqrt{T})$ rate of convergence, and the query complexity of the proposed black-box attack algorithm is linear in data dimension $d$. Finally, our empirical study on attacking both ImageNet dataset and MNIST dataset yield the best distortion in white-box setting and highest attack success rate/query complexity in black-box setting.

It would also be interesting to see the whether the performance of our Frank-Wolfe adversarial framework can be further improved by incorporating the idea of gradient/data priors \citep{ilyas2018prior}. We leave it as a future work.

%\newpage
\appendix

%\appendix
\section{Derivation of LMO for General $p \geq 1 $}\label{sec:lmo_dev}

The derivation of LMO is as follows:
denote $\hb = (\xb -\xb_{\ori})/\epsilon$, we have
\begin{align*}
    \argmin_{\|\xb - \xb_{\ori}\|_p \leq \epsilon } \langle \xb,  \mb_t \rangle  &=  \argmin_{\|\hb\|_p \leq 1} \epsilon \cdot\langle \hb, \mb_t \rangle   \\
    &=  \argmax_{\|\hb\|_p \leq 1} \epsilon \cdot\langle \hb, - \mb_t \rangle.
\end{align*}
By H\"older's inequality, the maximum value is reached when 
\begin{align*}
    |h_i| = c \cdot |(m_t)_i |^{\frac{1}{p-1}},
\end{align*}
where the subscript $i$ denotes the $i$-th element in the vector and $c$ is a universal constant. Together with the constraint of $\|\hb\|_p \leq 1$ we obtain
\begin{align*}
    h_i = - \frac{\sign\big((m_t)_i \big) \cdot |(m_t)_i |^{\frac{1}{p-1}}}{\Big( \sum_{i=1}^{d} |(m_t)_i |^{\frac{p}{p-1}}   \Big)^{\frac{1}{p}}}
\end{align*}
Therefore, we have
\begin{align*}
    x_i = - \epsilon \cdot \frac{\sign\big((m_t)_i \big) \cdot |(m_t)_i |^{\frac{1}{p-1}}}{\Big( \sum_{i=1}^{d} |(m_t)_i |^{\frac{p}{p-1}}   \Big)^{\frac{1}{p}}} + (x_{\ori})_i
\end{align*}
Submit the above result back into LMO solution,
% Since $\max_{\|\hb\|_p \leq 1} \langle \hb, -\nabla f(\xb_t) \rangle = \|\nabla f(\xb_t)\|_{p*}$, where $\|\cdot\|_{p*}$ denotes the dual norm of $\|\cdot\|_p$. We submit the LMO solution $\vb_t$ in the above formula,
for $p =2$ case, we have
% \begin{align*}
%     \langle (\vb_t-\xb_{\ori})/(\epsilon), -\nabla f(\xb_t) \rangle =  \|\nabla f(\xb_t)\|_{2}.
% \end{align*}
% It immediately implies that
\begin{align*}
    \vb_t = -\frac{\epsilon \cdot \mb_t}{\|\mb_t\|_2}  + \xb_{\ori}.
\end{align*}
For $p =\infty$ case, we have
% \begin{align*}
%     &\langle (\vb_t-\xb_{\ori})/(\epsilon), -\nabla f(\xb_t) \rangle =  \|\nabla f(\xb_t)\|_{1}.
% \end{align*}
% It immediately implies that
\begin{align*}
    \vb_t =  - \epsilon \cdot \sign( \mb_t) + \xb_{\ori}.
\end{align*}

\section{Proof of the Main Results}\label{sec:proof}
In the section, we provide the proofs of the technical lemmas and theorems in the main paper.

\subsection{Proof of Lemma \ref{lemma:m_t}}
\begin{proof}
By definition, we have
\begin{align}\label{eq:t1}
     \|\nabla f(\xb_t) - \mb_t\|_2 &= \| \nabla f(\xb_t) - \beta \mb_{t-1} - (1 - \beta) \nabla f(\xb_t)\|_2 \notag\\
     &= \beta \cdot \| \nabla f(\xb_t) - \mb_{t-1}\|_2 \notag\\
     &= \beta \cdot \| \nabla f(\xb_t) -\nabla f(\xb_{t-1}) + \nabla f(\xb_{t-1}) -  \mb_{t-1}\|_2 \notag\\
     &\leq \beta \cdot \| \nabla f(\xb_t) -\nabla f(\xb_{t-1}) \|_2 + \beta \cdot \|\nabla f(\xb_{t-1}) - \mb_{t-1}  \|_2 \notag\\
     &\leq \beta L \cdot \| \xb_t - \xb_{t-1}\|_2 + \beta \cdot \|\nabla f(\xb_{t-1}) - \mb_{t-1} \|_2,
\end{align}
where the first inequality follows triangle inequality and the second inequality holds due to Assumption \ref{assump:smooth}. In addition, we have the following result:
\begin{align}
    \|\xb_t - \xb_{t-1}\|_2 & = \gamma \|\mathbf{d}_{t-1}\|_2 \notag\\
    & = \gamma\|\vb_{t-1} - \xb_{t-1}\|_2\notag\\
     &\leq \gamma D. \label{eq:t2}
\end{align}
Substituting \eqref{eq:t2} into \eqref{eq:t1}, and recursively applying the above inequality yields that 
\begin{align*}
     \|\nabla f(\xb_t) - \mb_t\|_2 &\leq \gamma\big(\beta LD + \beta^2 LD + \ldots + \beta^t \| \nabla f(\xb_{0}) - \mb_{0} \|_2\big) \\
     & = \gamma\big(\beta LD + \beta^2 LD + \ldots + \beta^{t-1} LD\big) \\
     &\leq \frac{\gamma LD}{1 - \beta}.
\end{align*}

\end{proof}

\subsection{Proof of Theorem \ref{theorem:white}}
\begin{proof}
% The proof is similar to \cite{lacoste2016convergence}, hence we omit it. Note that the main difference is that \cite{lacoste2016convergence} uses time varying step size, while we use fixed step size.
First by Assumption \ref{assump:smooth}, we have
\begin{align}\label{eq:t3}
    f(\xb_{t+1}) &\leq f(\xb_t) + \nabla f(\xb_t)^\top (\xb_{t+1} - \xb_t) + \frac{L}{2}\|\xb_{t+1} -\xb_t\|_2^2 \notag\\
    &= f(\xb_t) + \gamma \nabla f(\xb_t)^\top (\vb_t - \xb_t) + \frac{L\gamma^2}{2}\|\vb_t - \xb_t\|_2^2 \notag\\
    &\leq f(\xb_t) + \gamma \nabla f(\xb_t)^\top (\vb_t - \xb_t) + \frac{LD^2\gamma^2}{2} \notag\\
    &= f(\xb_t) + \gamma \mb_t^\top (\vb_t - \xb_t)  + \gamma (\nabla f(\xb_t) - \mb_t)^\top (\vb_t - \xb_t) + \frac{LD^2\gamma^2}{2},
\end{align}
where the last inequality uses the bounded domain condition in Assumption \ref{assump:bounded_domain}.
Now define an auxiliary quantity:
\begin{align*}
    \hat \vb_t = \argmin_{\vb \in \cX} \langle \vb, \nabla f(\xb_t) \rangle.
\end{align*}
According to the definition of $g(\xb_t)$, this immediately implies 
\begin{align}\label{eq:t5}
    g(\xb_t) = - \langle \hat \vb_t - \xb_t, \nabla f(\xb_t) \rangle.
\end{align}
On the other hand, according to Line \ref{op:lmo} in Algorithm \ref{alg:fw}, we have
\begin{align*}
    \vb_t = \argmin_{\vb \in \cX} \langle \vb, \mb_t \rangle,
\end{align*}
which implies that
\begin{align}\label{eq:t4}
    \langle \vb_t, \mb_t \rangle \leq \langle \hat\vb_t, \mb_t \rangle.
\end{align}
Combining \eqref{eq:t3} and \eqref{eq:t4}, we further have
\begin{align}\label{eq:t7}
    f(\xb_{t+1}) 
    &\leq f(\xb_t) + \gamma \mb_t^\top (\hat \vb_t - \xb_t) + \gamma (\nabla f(\xb_t) - \mb_t)^\top (\vb_t - \xb_t) + \frac{LD^2\gamma^2}{2} \notag\\
    &= f(\xb_t) + \gamma \nabla f(\xb_t)^\top (\hat \vb_t - \xb_t) + \gamma (\nabla f(\xb_t) - \mb_t)^\top (\vb_t - 
    \hat \vb_t) + \frac{LD^2\gamma^2}{2}\notag\\
    &= f(\xb_t) - \gamma g(\xb_t) + \gamma (\nabla f(\xb_t) - \mb_t)^\top (\vb_t - 
    \hat \vb_t) + \frac{LD^2\gamma^2}{2}\notag\\
    &\leq f(\xb_t) - \gamma g(\xb_t) + \gamma D \cdot \|\nabla f(\xb_t) - \mb_t\|_2 + \frac{LD^2\gamma^2}{2},
\end{align}
where the first equality is obtained by rearranging the second and the third term on the right hand side of the first inequality, the second equality follows the definition of the Frank-Wolfe gap $g(\xb_t)$ \eqref{eq:t5}, and the last inequality holds due to Cauchy-Schwarz inequality.
By Lemma \ref{lemma:m_t}, we have 
\begin{align}\label{eq:t6}
     \|\nabla f(\xb_t) - \mb_t\|_2  
     &\leq \frac{\gamma LD}{1 - \beta}.
\end{align}
Substituting \eqref{eq:t6} into \eqref{eq:t7}, we obtain
% \begin{align*}
%     f(\xb_{t+1}) &\leq f(\xb_t) - \gamma g(\xb_t) + \frac{LD^2\gamma}{1 - \beta} + \frac{LD^2\gamma^2}{2}.
% \end{align*}
\begin{align*}
    f(\xb_{t+1}) &\leq f(\xb_t) - \gamma g(\xb_t) + \frac{LD^2\gamma^2}{1 - \beta} + \frac{LD^2\gamma^2}{2}.
\end{align*}
Telescoping over $t=0,\ldots,T-1$ of the above inequality, we obtain
\begin{align}\label{eq:t8}
    f(\xb_{T}) &\leq f(\xb_0) -  \sum_{t=0}^{T-1} \gamma g(\xb_t) + \frac{TLD^2\gamma^2}{1 - \beta} +\frac{TLD^2\gamma^2}{2}.
    % \\
    % &\leq f(\xb_0) -  \gamma T\tilde g_T + \frac{ TLD^2\gamma^2}{1 - \beta} +  \frac{TLD^2\gamma^2}{2},
\end{align}
Rearranging \eqref{eq:t8} yields
\begin{align}\label{eq:t9}
    \frac{1}{T}\sum_{t=0}^{T-1} g(\xb_t) &\leq \frac{f(\xb_0) - f(\xb_{T})}{T\gamma} + \gamma \bigg(\frac{LD^2}{1-\beta} + \frac{LD^2}{2}\bigg)\notag \\
    &\leq \frac{f(\xb_0) - f(\xb^*)}{T\gamma} + \gamma \bigg(\frac{LD^2}{1-\beta} + \frac{LD^2}{2}\bigg),
\end{align}
where the second inequality is due to the optimality that $f(\xb_{T}) \geq f(\xb^*)$. 
\eqref{eq:t9} further implies
%where the second inequality follows from the definition of
\begin{align*}
    \tilde g_T = \min_{1 \leq k \leq T} g(\xb_k)
    &\leq \frac{f(\xb_0) - f(\xb^*)}{ T\gamma} + \gamma\bigg(\frac{ LD^2}{1 - \beta} + \frac{LD^2}{2}\bigg).
    % \\
    % &\leq 2\sqrt{ \frac{LD^2 C_\beta  (f(\xb_0) - f(\xb^*))}{ T}} ,
\end{align*}
Let $\gamma = \sqrt{ 2(f(\xb_0) - f(\xb^*))/( C_\beta LD^2T)}$ where $C_\beta = (3-\beta)/(1-\beta)$, we have
\begin{align*}
    \tilde g_T  
    &\leq \sqrt{ \frac{2LD^2 C_\beta  (f(\xb_0) - f(\xb^*))}{ T}}.
\end{align*}
\end{proof}

\subsection{Proof of Lemma \ref{lemma:zero_order}}
\begin{proof} 
% The proof for Lemma \ref{lemma:zero_order} follows from the proof of Lemma 4.1 in \cite{gao2018information} and Lemma 10 in \cite{shamir2017optimal} and hence we omit it.
% For simplicity we denote $f(\cdot)$ by $f(\cdot)$ for the rest of the proof.
First, we have
\begin{align}\label{eq:l0}
    \EE \|\nabla f(\xb) - \qb\|_2 &\leq \|\nabla f(\xb) - \EE [\qb] \|_2 +  \EE \|\qb - \EE [\qb]\|_2\notag\\
    &\leq \underbrace{\|\nabla f(\xb) - \EE [\qb] \|_2}_{I_1} +  \underbrace{\sqrt{\EE \|\qb - \EE [\qb] \|_2^2}}_{I_2},
\end{align}
where the first inequality follows from triangle inequality, the second inequality holds due to Jensen's inequality.
Let us denote $\bpsi_i = \frac{d}{2\delta b} \big(  f(\xb + \delta \ub_i) - f(\xb - \delta \ub_i) \big) \ub_i$. For term $I_1$, we have
\begin{align*}
    \EE [ \bpsi_i ] &= \EE_\ub \bigg[ \frac{d}{2\delta b} \big( f(\xb + \delta \ub_i) - f(\xb - \delta \ub_i) \big) \ub_i \bigg] \\
    &= \EE_\ub \bigg[ \frac{d}{2\delta b}    f(\xb + \delta \ub_i)   \ub_i \bigg]  + \EE_\ub \bigg[ \frac{d}{2\delta b}  f(\xb - \delta \ub_i) (-  \ub_i) \bigg]\\
    &= \EE_\ub \bigg[ \frac{d}{\delta b}  f(\xb + \delta \ub_i)  \ub_i \bigg]\\
    &= \frac{1}{b} \nabla f_\delta(\xb  ),
\end{align*}
where the third equality holds due to symmetric property of $\ub_i$ and the last equality follows from Lemma \ref{lemma:aux}. Therefore, we have
\begin{align*} 
    \EE [ \qb ] &= \EE \bigg[\sum_{i = 1}^b \bpsi_i \bigg] = \nabla f_\delta(\xb  ). 
\end{align*}
This further implies that 
\begin{align}\label{eq:l1}
    \|\nabla f(\xb) - \EE [\qb] \|_2 = \|\nabla f(\xb) - \nabla f_\delta(\xb) \|_2 \leq \frac{\delta dL}{2},
\end{align}
where the inequality follows from Lemma \ref{lemma:aux}.

For term $I_2$, note that $\bpsi_i$'s are independent from each other due to the independence of $\ub_i$, we have
\begin{align}\label{eq:l2}
    \EE \|\qb  - \EE[\qb]\|_2^2 &= \EE \bigg\|  \sum_{i = 1}^b \Big[  \bpsi_i - \EE \bpsi_i \Big] \bigg\|_2^2 \notag\\
    &=  \sum_{i = 1}^b \EE   \big\|  \bpsi_i - \EE \bpsi_i \big\|^2 \leq \sum_{i = 1}^b \EE   \big\|  \bpsi_i \big\|^2. 
\end{align}
Note that for term $\EE \big\|\bpsi_i \big\|^2$, we have
\begin{align}\label{eq:l3}
    \EE \big\|  \bpsi_i \big\|^2 
    &= \EE_\ub \bigg\| \frac{d}{2\delta b} \big( f(\xb + \delta \ub_i ) - f(\xb) + f(\xb) - f(\xb - \delta \ub_i ) \big) \ub_i \bigg\|_2^2\notag\\
    &\leq \frac{1}{2b^2} \EE_\ub \bigg\| \frac{d}{\delta} \big( f(\xb + \delta \ub_i ) - f(\xb)  \big) \ub_i \bigg\|_2^2  + \frac{1}{2b^2} \EE_\ub \bigg\| \frac{d}{\delta} \big( f(\xb  ) - f(\xb -  \delta \ub_i)  \big) \ub_i \bigg\|_2^2\notag\\
    &= \frac{1}{b^2} \EE_\ub \bigg\| \frac{d}{\delta} \big( f(\xb + \delta \ub_i ) - f(\xb)  \big) \ub_i \bigg\|_2^2\notag\\
    &\leq \frac{1}{b^2} \bigg( 2d\|\nabla f(\xb)\|_2^2 + \frac{1}{2}\delta^2 L^2 d^2 \bigg),
\end{align}
where the first inequality is due to the fact that $(a + b)^2 \leq 2a^2 + 2b^2$, the second equality follows from the symmetric property of $\ub_i$ and the last inequality is by Lemma \ref{lemma:aux}.
Also note that by Assumption \ref{assump:smooth} and \ref{assump:grad_zero_bound} we have
\begin{align*}
    \|\nabla f(\xb)\|_2^2 \leq (\|\nabla f(\zero))\|_2 + L\|\xb\|_2)^2 \leq (G+ LD)^2.
\end{align*}
Therefore, \eqref{eq:l3} can be further written as 
\begin{align}\label{eq:l4}
    \EE \big\|  \bpsi_i \big\|^2 
    &\leq \frac{1}{b^2} \bigg( 2d(G+ LD)^2 + \frac{1}{2}\delta^2 L^2 d^2 \bigg).
\end{align}
Substituting \eqref{eq:l4} into \eqref{eq:l2} we have 
\begin{align}\label{eq:l6}
    \EE \|\qb  - \EE[\qb]\|_2^2 \leq  \frac{1}{b}  \bigg( 2d(G+ LD)^2 + \frac{1}{2}\delta^2 L^2 d^2 \bigg).
\end{align}
Combining \eqref{eq:l0}, \eqref{eq:l1} and \eqref{eq:l6} we obtain
\begin{align*} 
    \EE \|\nabla f(\xb) - \qb\|_2 &\leq \frac{\delta Ld}{2} + \frac{1}{b}  \bigg( 2d(G+ LD)^2 + \frac{1}{2}\delta^2 L^2 d^2 \bigg) \\
    &\leq \frac{\delta Ld}{2} + \frac{ 2\sqrt{d}(G+ LD) +  \delta L d }{\sqrt{2b}}  .
\end{align*}

\end{proof}

\subsection{Proof of Theorem \ref{theorem:black}}
\begin{proof}
First by Assumption \ref{assump:smooth}, we have
\begin{align}\label{eq:tb1}
    f(\xb_{t+1}) &\leq f(\xb_t) + \nabla f(\xb_t)^\top (\xb_{t+1} - \xb_t) + \frac{L}{2}\|\xb_{t+1} -\xb_t\|_2^2 \notag\\
    &= f(\xb_t) + \gamma \nabla f(\xb_t)^\top (\vb_t - \xb_t) + \frac{L\gamma^2}{2}\|\vb_t - \xb_t\|_2^2 \notag\\
    &\leq f(\xb_t) + \gamma \nabla f(\xb_t)^\top (\vb_t - \xb_t) + \frac{LD^2\gamma^2}{2}\notag\\
    &= f(\xb_t) + \gamma \mb_t^\top (\vb_t - \xb_t)  + \gamma (\nabla f(\xb_t) - \mb_t)^\top (\vb_t - \xb_t) + \frac{LD^2\gamma^2}{2},
\end{align}
where the second inequality uses the bounded domain condition in Assumption \ref{assump:bounded_domain}.
Now define an auxiliary quantity:
\begin{align*}
    \hat \vb_t = \argmin_{\vb \in \cX} \langle \vb, \nabla f(\xb_t) \rangle.
\end{align*}
According to the definition of $g(\xb_t)$, this immediately implies 
\begin{align}\label{eq:tb2}
    g(\xb_t) = - \langle \hat \vb_t - \xb_t, \nabla f(\xb_t) \rangle.
\end{align}
On the other hand, according to Line \ref{op:lmo} in Algorithm \ref{alg:fw}, we have
\begin{align*}
    \vb_t = \argmin_{\vb \in \cX} \langle \vb, \mb_t \rangle,
\end{align*}
which implies that
\begin{align}\label{eq:tb3}
    \langle \vb_t, \mb_t \rangle \leq \langle \hat\vb_t, \mb_t \rangle.
\end{align}
Combining \eqref{eq:tb1} and \eqref{eq:tb3}, we further have
\begin{align}\label{eq:tb4}
    f(\xb_{t+1}) 
    &\leq f(\xb_t) + \gamma \mb_t^\top (\hat \vb_t - \xb_t) + \gamma (\nabla f(\xb_t) - \mb_t)^\top (\vb_t - \xb_t) + \frac{LD^2\gamma^2}{2}\notag\\
    &= f(\xb_t) + \gamma \nabla f(\xb_t)^\top (\hat \vb_t - \xb_t) + \gamma (\nabla f(\xb_t) - \mb_t)^\top (\vb_t - 
    \hat \vb_t) + \frac{LD^2\gamma^2}{2}\notag\\
    &= f(\xb_t) - \gamma g(\xb_t) + \gamma (\nabla f(\xb_t) - \mb_t)^\top (\vb_t - 
    \hat \vb_t) + \frac{LD^2\gamma^2}{2}\notag\\
    &\leq f(\xb_t) - \gamma g(\xb_t) + \gamma D \cdot \|\nabla f(\xb_t) - \mb_t\|_2 + \frac{LD^2\gamma^2}{2},
\end{align}
where the first equality is obtained by rearranging the second and the third term on the right hand side of the first inequality, the second equality follows the definition of the Frank-Wolfe gap $g(\xb_t)$ \eqref{eq:tb2}, and the last inequality holds due to Cauchy-Schwarz inequality.

% By Lemma \ref{lemma:m_t}, we have 
% \begin{align}\label{eq:t6}
%      \|\nabla f(\xb_t) - \mb_t\|_2  
%      &\leq \frac{\gamma LD}{1 - \beta}.
% \end{align}
% Substituting \eqref{eq:t6} into \eqref{eq:t7}, we obtain

Take expectations for both sides of \eqref{eq:tb4}, we have
\begin{align}\label{eq:tb5}
    \EE [f(\xb_{t+1})] 
    &\leq \EE [f(\xb_t)] - \gamma \EE [g(\xb_t)] + \gamma D \cdot \EE \|\nabla f(\xb_t) - \mb_t\|_2 + \frac{LD^2\gamma^2}{2}\notag\\
    &\leq \EE [f(\xb_t)] - \gamma \EE [g(\xb_t)] + \gamma D \cdot \big(\EE\| \qb_t - \mb_t \|_2 + \|\nabla f(\xb_t)  - \qb_t\|_2   \big) + \frac{LD^2\gamma^2}{2} \notag\\
    &\leq \EE [f(\xb_t)] - \gamma \EE [g(\xb_t)] + \gamma D \Bigg( \EE\| \qb_t - \mb_t\|_2 + \frac{\delta Ld}{2} + \frac{ 2\sqrt{d}(G+ LD) +  \delta L d }{\sqrt{2b}} \Bigg) + \frac{LD^2\gamma^2}{2} ,
\end{align}
where the second inequality follows from triangle inequality and the third inequality holds by Lemma \ref{lemma:zero_order}.
%From Lemma \ref{lemma:m_t}, we already know $\|\xb_t - \xb_{t-1}\|_2 \leq \gamma D$. 
Similar to the proof of Lemma \ref{lemma:m_t}, we have
\begin{align}\label{eq:tb6}
     \EE \|\qb_t - \mb_t\|_2  
     &\leq \beta \cdot \EE\|\qb_t - \qb_{t-1}\|_2 + \beta \cdot \EE\|\qb_{t-1} - \mb_{t-1}  \|_2 \notag\\
     &\leq \beta \big( \EE\|\nabla f(\xb_t) - \qb_t \|_2  + \EE\|\nabla f(\xb_{t-1})- \qb_{t-1}\|_2  + \EE\|\nabla f(\xb_t) - \nabla f(\xb_{t-1})\|_2 \big)+ \beta \cdot \EE\| \qb_{t-1} - \mb_{t-1} \|_2\notag\\
     &\leq \beta \bigg(\delta Ld  + \frac{ 2\sqrt{2d}(G+ LD) +  \sqrt{2} \delta L d }{\sqrt{b}} +  L\|\xb_t -\xb_{t-1}\|_2 \bigg)  + \beta \cdot \EE\| \qb_{t-1} - \mb_{t-1} \|_2\notag\\
     &\leq \beta \bigg(\delta Ld  + \frac{ 2\sqrt{2d}(G+ LD) +  \sqrt{2} \delta L d }{\sqrt{b}} + \gamma LD \bigg)  + \beta \cdot \EE\| \qb_{t-1} - \mb_{t-1} \|_2,
\end{align}
where the first and the second inequalities follow from triangle inequality, the third inequality follows from Lemma \ref{lemma:zero_order}, and the last inequality follows from \eqref{eq:t2}. 
%Similar to Lemma \ref{lemma:m_t},
Recursively applying \eqref{eq:tb6} yields that 
\begin{align}\label{eq:tb7}
     \EE \|\qb_t - \mb_t\|_2  
     &\leq \frac{1}{1-\beta} \bigg(\delta Ld  + \frac{ 2\sqrt{2d}(G+ LD) +  \sqrt{2} \delta L d }{\sqrt{b}} + \gamma LD \bigg).
\end{align}
Combining \eqref{eq:tb5} and \eqref{eq:tb7}, we have
\begin{align}\label{eq:tb8}
    \EE [f(\xb_{t+1})] 
    &\leq \EE [f(\xb_t)] - \gamma \EE [g(\xb_t)] + \frac{LD^2\gamma^2}{2} + \gamma D \cdot \Bigg(\frac{3-\beta}{1-\beta} \bigg(\frac{\delta Ld}{2} + \frac{ 2\sqrt{d}(G+ LD) +  \delta L d }{\sqrt{2b}}  \bigg) + \frac{\gamma LD}{1 - \beta}  \Bigg).
\end{align}
Telescoping over $t$ of \eqref{eq:tb8}, we obtain
\begin{align}\label{eq:tb9}
    \EE [f(\xb_{T})] 
    &\leq f(\xb_0) -  \sum_{t=0}^{T-1} \gamma \EE[g(\xb_t)] + \frac{TLD^2\gamma^2}{2} + \gamma DT \Bigg(\frac{3-\beta}{1-\beta} \bigg(\frac{\delta Ld}{2} + \frac{ 2\sqrt{d}(G+ LD) +  \delta L d }{\sqrt{2b}}  \bigg) + \frac{\gamma LD}{1 - \beta}  \Bigg).
    % &\leq f(\xb_0) -  \gamma T g_a  + \frac{TLD^2\gamma^2}{2} +\gamma DT  \Bigg( \frac{\delta Ld}{2} + \frac{ 2\sqrt{d}(G+ LD) +  \delta L d }{\sqrt{2b}}  \Bigg),
\end{align}
Rearranging \eqref{eq:tb9} yields
\begin{align}\label{eq:tb10}
    \frac{1}{T}\sum_{t=0}^{T-1}\EE[g(\xb_t)] &\leq \frac{f(\xb_0) - \EE [f(\xb_{T})] }{T\gamma} +  D \cdot \frac{3-\beta}{1-\beta} \bigg( \gamma LD +  \frac{\delta Ld}{2} + \frac{ 2\sqrt{d}(G+ LD) +  \delta L d }{\sqrt{2b}}  \bigg)\notag \\
    &\leq \frac{f(\xb_0) - f(\xb^*)}{T\gamma} + D \cdot \frac{3-\beta}{1-\beta} \bigg( \gamma LD +  \frac{\delta Ld}{2} + \frac{ 2\sqrt{d}(G+ LD) +  \delta L d }{\sqrt{2b}}  \bigg),
\end{align}
where the second inequality is due to the optimality that $f(\xb_{T}) \geq f(\xb^*)$. 
By the definition of $\tilde g_T$ in Algorithm 
\ref{alg:fw-blackbox}, we further have
\begin{align*}
    \EE [\tilde g_T] 
    &\leq \frac{f(\xb_0) - f(\xb^*)}{ T\gamma} + D \cdot \frac{3-\beta}{1-\beta} \bigg( \gamma LD +  \frac{\delta Ld}{2} + \frac{ 2\sqrt{d}(G+ LD) +  \delta L d }{\sqrt{2b}}  \bigg).
\end{align*}
Let $\gamma = \sqrt{(f(\xb_0) - f(\xb^*))/( C_\beta LD^2T)}$ where $C_\beta = (3-\beta)/(1-\beta)$, $b = Td$ and $\delta = \sqrt{1/(Td^2)}$, we have
\begin{align*}
   \EE [\tilde g_T]  &\leq \frac{f(\xb_0) - f(\xb^*)}{ T\gamma} + C_\beta LD^2\gamma + C_\beta D  \Bigg( \frac{\delta Ld}{2} + \frac{ 2\sqrt{d}(G+ LD) +  \delta L d }{\sqrt{2b}}  \Bigg) \\
    &\leq \frac{D}{ \sqrt{T}} \Big ( 2\sqrt{C_\beta L(f(\xb_0) - f(\xb^*))} + C_\beta (L +G+ LD)  \Big).
\end{align*}
\end{proof}

% \section{Proof of the Technical Lemmas}
% In the section, we provide the proof of the key technical lemmas mentioned in the paper.

\section{Auxiliary Lemma}
\begin{lemma}[Lemma 4.1 in \cite{gao2018information}]
\label{lemma:aux}
Under Assumption \ref{assump:smooth}, let $f_\delta(\xb) = \EE_\ub[f(\xb + \delta \ub) ]$ where $\ub$ is sampled uniformly from the Euclidean unit sphere with with $\|\ub\|_2 = 1$, we have
\begin{align*}
    &\nabla f_\delta(\xb) = \EE_\ub \bigg[ \frac{d}{\delta}  f(\xb + \delta \ub) \ub \bigg],\\
    &\| \nabla f(\xb) - \nabla f_\delta(\xb)\|_2 \leq \frac{\delta dL}{2},\\
    & \EE_\ub \bigg\| \frac{d}{\delta} \big( f(\xb + \delta  \ub  ) - f(\xb )  \big) \ub  \bigg\|_2^2 = \bigg( 2d\|\nabla f(\xb )\|_2^2 + \frac{1}{2}\delta^2 L^2 d^2 \bigg),
\end{align*}
where $L$ is the smoothness parameter and $d$ is the data dimension.
\end{lemma}

\section{Additional Experimental Details}\label{sec:exp_setting}

\subsection{Parameter Settings}

As we mentioned in the main paper, we performed grid search to tune the hyper-parameters for all algorithm to ensure a fair comparison.  
	In detail, in white-box experiments, 
	for Algorithm \ref{alg:fw}, we tune the step size $\gamma_t$ by searching the grid $\{0.1, 0.2, \ldots, 0.9\}$ on both datasets. For PGD and MI-FGSM, we tune the learning rate by searching the grid $\{0.05, 0.10, \ldots, 0.25, 0.30\}$ on MNIST and $\{0.01, 0.02, \ldots, 0.4, 0.5]$ on ImageNet. For both FW and MI-FGSM, we tune the momentum parameter $\beta$ by searching the grid $\{0.1, 0.5, 0.9, 0.99\}$. 
	In black-box experiments, for Algorithm \ref{alg:fw-blackbox}, we tune the step size $\gamma_t$ by the grid $\{0.1/\sqrt{t}, 0.2/\sqrt{t}, \ldots, 0.9/\sqrt{t}\}$, the momentum parameter $\beta$ by the grid $\{0.5, 0.9, 0.99, 0.999, 0.9999\}$, the sample size for gradient estimation $b$ by the grid $\{5,10,\ldots, 50\}$, and the sampling parameter $\delta$ by the grid $\{0.1, 0.01, 0.001, 0.0001\}$.
	For NES-PGD and Bandit, we tune the learning rate by the grid $\{0.01, 0.02, \ldots, 0.10\}$ on MNIST and $\{0.001, 0.005, 0.01, \ldots, 0.02\}$ on ImageNet.  
	For other parameters, we follow the recommended parameter setting in their original papers.
 We report the hyperparameters tuned by grid search in the sequel. In detail, for white-box experiments, we list the hyper-parameters used for all algorithms in Table \ref{table:para_white}.
For black-box experiments, we list the hyper-parameters for  Frank-Wolfe black-box attack algorithm in Table \ref{table:para_black_fw}. We also list the hyper-parameters used for NES-PGD and Bandit black-box attack algorithms in Tables \ref{table:para_black_nes} and \ref{table:para_black_bandit} respectively.

\begin{table}[h!]
  \caption{Parameters used in targeted white-box experiments.
  }
  \label{table:para_white}
  \begin{center}
  \begin{small}
  \begin{tabular}{l|ccc|ccc}
    \toprule
    \multicolumn{1}{c}{\multirow{2}{*}{Parameters}} & \multicolumn{3}{c}{MNIST} &
    \multicolumn{3}{c}{ImageNet}\\
    % \cline{2-5}
    \cmidrule(r){2-4}
    \cmidrule(r){5-7}
    \multicolumn{1}{c}{} & PGD & MI-FGSM & FW & PGD & MI-FGSM & FW \\
    \midrule 
    $\epsilon$      & \multicolumn{3}{c|}{ 0.3 }& \multicolumn{3}{c}{ 0.05 } \\
    $\gamma_t$  & 0.1 & 0.1 & 0.5 & 0.03 & 0.03 & 0.1  \\
    $\beta$      & - & 0.9 & 0.9 & - & 0.9 & 0.9  \\
    \bottomrule
  \end{tabular}
  \end{small}
  \end{center}
\end{table}

\begin{table}[h!]
\caption{Parameters used in Frank-Wolfe black-box experiments.}
\label{table:para_black_fw}
\begin{center}
\begin{small}
\begin{sc}
\begin{tabular}{lcc}
\toprule
{Parameter} & {MNIST} & {ImageNet} \\
\midrule
$\{\gamma_t\}$  & $0.8/\sqrt{t}$ & $0.8/\sqrt{t}$\\
$\beta$         & 0.99     & 0.999\\
$b$             & 25        & 25\\
$\delta$        & 0.01      & 0.01\\
\bottomrule
\end{tabular}
\end{sc}
\end{small}
\end{center}
\end{table}

\begin{table}[h!]
\caption{Parameters used in NES-PGD black-box experiments.}
\label{table:para_black_nes}
\begin{center}
\begin{small}
\begin{sc}
\begin{tabular}{lcc}
\toprule
{Parameter} & {MNIST} & {ImageNet} \\
\midrule
learning rate  & $0.02$ & $0.005$ \\
number of finite difference estimations per step  & 25        & 25\\
finite difference probe  & 0.001      & 0.001\\
\bottomrule
\end{tabular}
\end{sc}
\end{small}
\end{center}
\end{table}

\begin{table}[h!]
\caption{Parameters used in Bandit black-box experiments.}
\label{table:para_black_bandit}
\begin{center}
\begin{small}
\begin{sc}
\begin{tabular}{lcc}
\toprule
{Parameter} & {MNIST} & {ImageNet} \\
\midrule
learning rate   & $0.03$ & $0.01$ \\
% number of finite difference estimations per step  & 1       & 1\\
finite difference probe  & 0.1      & 0.1\\
online convex optimization learning rate & 0.001 & 0.0001\\
data-dependent prior size & 8 & 50\\
bandit exploration & 0.01 & 0.01\\
\bottomrule
\end{tabular}
\end{sc}
\end{small}
\end{center}
\end{table}

\bibliography{adv}
\bibliographystyle{ims}

\end{document}